\documentclass[conference]{IEEEtran}
\IEEEoverridecommandlockouts
\usepackage{cite}
\usepackage{amsmath,amssymb,amsfonts}
\usepackage{mathrsfs, cases, algorithm}
\usepackage{graphicx}
\usepackage{textcomp}
\usepackage{xcolor, hyperref}
\usepackage[noend]{algpseudocode}

\newcommand{\R}{\mathbb{R}}

\def\BibTeX{{\rm B\kern-.05em{\sc i\kern-.025em b}\kern-.08em
    T\kern-.1667em\lower.7ex\hbox{E}\kern-.125emX}}
\begin{document}

\title{NENET: An Edge Learnable Network for Link Prediction in Scene Text}

\author{\IEEEauthorblockN{1\textsuperscript{st} Mayank Kumar Singh}
\IEEEauthorblockA{\textit{Electrical Engineering Department} \\
\textit{Indian Institute of Technology, Bombay}\\
Mumbai, India \\
mayanksingh@iitb.ac.in}
\and
\IEEEauthorblockN{2\textsuperscript{nd} Sayan Banerjee}
\IEEEauthorblockA{\textit{Electrical Engineering Department} \\
\textit{Indian Institute of Technology, Bombay}\\
Mumbai, India \\
154070021@iitb.ac.in}
\and
\IEEEauthorblockN{3\textsuperscript{rd} Subhasis Chaudhuri}
\IEEEauthorblockA{\textit{Electrical Engineering Department} \\
\textit{Indian Institute of Technology, Bombay}\\
Mumbai, India \\
sc@ee.iitb.ac.in}
}
\maketitle

\begin{abstract}
  Text detection in scenes based on deep neural networks have shown promising results. Instead of using word bounding box regression, recent state-of-the-art methods have started focusing on character bounding box and pixel-level prediction. This necessitates the need to link adjacent characters, which we propose in this paper using a novel Graph Neural Network (GNN) architecture that allows us to learn both node and edge features as opposed to only the node features under the typical GNN. The main advantage of using GNN for link prediction lies in its ability to connect characters which are spatially separated and have an arbitrary orientation. We show our concept on the well known SynthText dataset, achieving top results as compared to state-of-the-art methods.
\end{abstract}


\begin{figure*}[ht]
  \begin{center}
      \includegraphics[width=\linewidth]{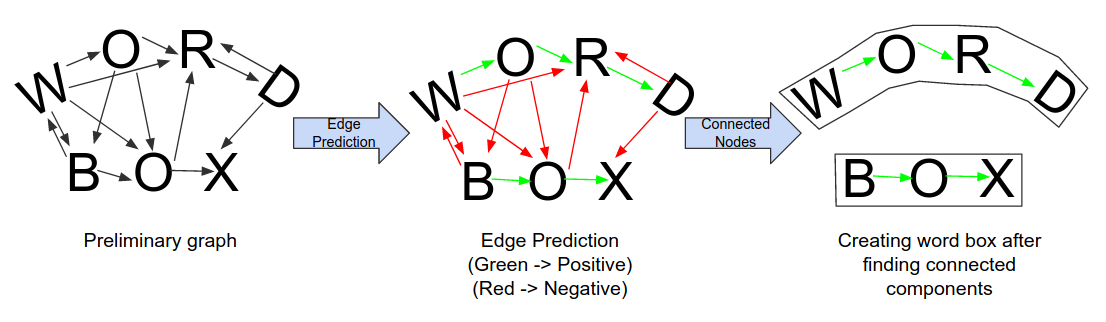}
  \end{center}
  \caption{Creating word boxes, given character boxes, using link prediction.}
  \label{fig:HowGCNNWorks}
\end{figure*}

\section{Introduction}

Recently, detecting and recognizing texts in natural scenes have been an area of prime interest, both in the industry as well as in the academic field. Extracting and understanding text from images have helped automate visa applications, loan processing systems, navigation and is a must for self-driven cars.

In the current literature, text detection has been treated separately to text recognition \cite{CRNN, StarNetRecognition, GRCNN}, while both are important for text extraction. In this paper, we focus on the more challenging scenario, which is the scene text detection. 

Scene text detection and extraction have seen a lot of advancements. Some of the earlier approaches, \cite{ICDAR2013, ETECNN, CTPN} predicted word bounding boxes by assuming the text to be horizontal and machine printed. As a result those methods perform sub-optimally on natural images since natural images contain a lot of oriented and irregularly shaped text. To address these limitations, \cite{EAST, PixelLink, SegLink, TextBoxes, MaskTextSpotter} relaxed the assumption by predicting oriented bounding boxes. Further relaxations were introduced in some recent methods which can predict irregularly shaped text as in \cite{TextSnake, CRAFT}.

The proposed method is inspired by \cite{CRAFT} in which we estimate the character bounding boxes but instead of linking them using affinity boxes, we construct a graph where each character is a node and we use a graph neural network for edge prediction to join two adjacent characters. Affinity heat maps computed in \cite{CRAFT} fails to join characters that are spatially separated, as well as characters that have a high aspect ratio, refer to Figure \ref{fig:ProblemsWithCRAFT} for illustration. The proposed framework with the help of a novel graph neural network operation can handle those aforementioned challenges efficiently.

Existing graph convolution methods \cite{EdgeConvolution, StarNet, PointNet, LinkPrediction} focus on training and propagating node features, where edges serve the purpose of defining the adjacency matrix. For example, part segmentation of point clouds, \cite{EdgeConvolution}, assumes fixed classes and each point/node predicts its class affiliation. This method restricts the number of unique objects to an upper threshold and does not apply to our task in which there can be a large number of words in a single image. To eliminate this problem, we propose to use link prediction which can effectively create any number of groups of characters. We also propose a novel method of using multi-dimension features for edges which are also trained and propagated and used to predict link between adjacent characters, see Figure \ref{fig:HowGCNNWorks} for illustration. We also show that our method performs better than existing graph convolution methods for the task of linking character bounding boxes to form words. 

Our main contributions are as follows:
\begin{itemize}
    \item We propose a novel GNN framework in which edge as well node features are learnt and propagated unlike existing methods which only use node features.
    \item The proposed GNN is applied in the spatial domain and works on variable adjacency matrices unlike existing works which focus on spectral methods.
    \item We propose a new method of linking estimated character boxes for text detection using GNN.
    \item We advance the current state-of-the-art method using the proposed technique.
     
\end{itemize}

Since the proposed network learns both Node and Edge parameters, we call it NENET.

\section{Related Work}

Much of the previous works\cite{EAST, TextBoxes, MaskTextSpotter} have got their motivation from \cite{YOLOV3, MaskRCNN} which were state of the art methods for object detection using bounding box regression. These methods require prior anchor boxes, which is a huge drawback for text detection as aspect ratio of word bounding boxes has a very high variance. Further, the assumption of words to be enclosed in boxes works poorly on curved and irregular shaped text \cite{TotalText, CTW1500}. Without using bounding box regression, instance segmentation of words was achieved by PixelLink\cite{PixelLink}, though the final output was still an oriented word bounding box. \cite{TextSnake} used a novel approach of segmenting the word center line along with a circle for each point on the line, allowing them to predict words of arbitrary shape. \cite{CRAFT} predicted character bounding boxes and formed word bounding boxes by linking them using affinity boxes.

Convolution in the non-Euclidean domain has drawn a lot of attention after the work of Bruna \textit{et al.} \cite{bruna2013spectral}. They formulated graphs into the spectral domain using eigenvectors of the corresponding graph Laplacian matrix and performed filtering in the spectral domain. Subsequently, the methods in \cite{defferrard2016convolutional,kipf2016semi,duvenaud2015convolutional} have proposed solutions to reduce the computational complexity of the operation of graph filtering and to improve filter performance. But all of the spectral methods demand the nature of input graphs to be homogeneous which is a strong constraint for real-life vision problems. To address that problem Monti \textit{et al.} \cite{monti2016geometric} proposed a novel path operator which can handle irregular local connectivity at different nodes. But to compute this operator at any node, it requires information of relative spatial distance and orientation of every other node with respect to the reference node \cite{monti2016geometric}. 
Graph CNN has been applied for different operations such as 3D shape segmentation \cite{yi2017syncspecCNN}, molecular fingerprint recognition \cite{duvenaud2015convolutional}, molecular quantum interaction \cite{schutt2017schnet}, action recognition \cite{huang2017deep}, point cloud segmentation \cite{EdgeConvolution}. But, with the best of our knowledge this is the first time graph CNN is used for word detection. Furthermore, unlike the existing graph neural network approaches, the proposed method propagates node and edge information simultaneously to learn better node as well as edge features.

\subsection{Most Similar Work}
\begin{figure*}[ht]
  
    \includegraphics[width=\textwidth]{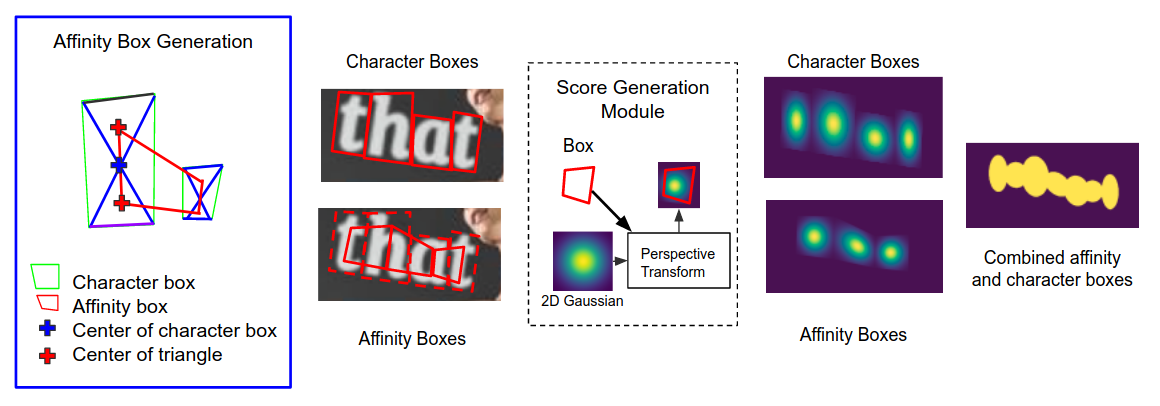}
    \caption{Illustration of ground truth generation in CRAFT algorithm. We use the same pipeline for character heat map generation.The figure is adapted from \cite{CRAFT}.}
    \label{fig:CRAFTPipeLine}
\end{figure*}

We build upon the work done in CRAFT\cite{CRAFT}. CRAFT predicts character regions and the affinity between characters. Refer to Figure \ref{fig:CRAFTPipeLine} for illustration. A fully convolutional network architecture based on VGG-16\cite{VGG16} is used for character and affinity prediction. The ground truth for the character is generated by pre-computing a 2-dimensional isotropic Gaussian map, computing the perspective transform between the Gaussian map region and each character box, and then warping the Gaussian isotropic map to the box area. 

For the generation of the affinity map which connects two adjacent characters, the centroid of the upper and lower triangle, formed by joining the diagonals of each character box, are connected. Similar to generating the ground truth for the character box, the affinity map is generated by warping a pre-computed isotropic Gaussian map.

During inference, the thresholded character and affinity map are combined and connected component labeling is performed on it. Lastly, an oriented minimum area rectangle is formed enclosing the connected components.

\begin{figure}[ht]
    \begin{center}
        \includegraphics[width=\linewidth]{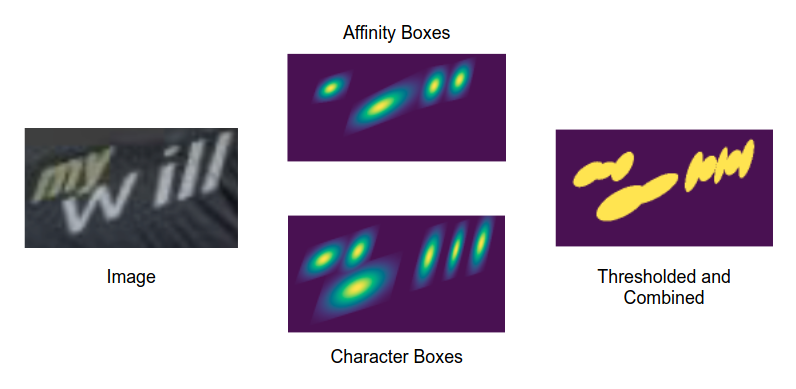}
    \end{center}
    \caption{Using the affinity box as mentioned in CRAFT creates problems as shown above. The target affinity box is not able to join the characters "w" and "i" which creates false labels and hinders training. Increasing the spread of the Gaussian map results in unwanted links. }
    \label{fig:ProblemsWithCRAFT}
\end{figure}

Some of the shortcomings of this approach are

\begin{itemize}
  \item The affinity box is not able to connect thin or spatially separated characters, which creates false target labels. Refer to Figure \ref{fig:ProblemsWithCRAFT} for illustration.
  \item The model has a fixed receptive field, and the predicted affinity box cannot connect characters which are spatially far apart but belong to the same word.
\end{itemize}

As we propose to use a GNN to predict links among characters, there are no false target labels. 
Spatially far apart characters are only separated by a small number of hops in the graph domain. Hence even if the receptive field of the NENET is small, it can link spatially secluded characters. 

\subsection{Graph Convolution}

Given a graph $\mathcal{G}=\{\mathcal{V},\mathcal{E}\}$, where $\mathcal{V}$ denotes the the collection of $n$ nodes and  $\mathcal{E}$ denotes the set of edges as $\mathcal{E}\subseteq \mathcal{V} \times \mathcal{V}$. With an input node features $X_{in}=[x_1^0,x_2^0,...,x_n^0]\in\mathbb{R}^{d\times n}$ and an adjacency matrix $A \in \R^{n \times n}$, vanilla graph convolution \cite{kipf2016semi} is performed as follows 
\begin{equation}
    X_1^\intercal = ReLU ((D^{-1/2}(I_n+A)D^{-1/2}) \times X_{in}^\intercal \times H^1)
    \label{eqn:existing_graphcnn_eqn}
\end{equation}
where $H^1 \in \R^{d \times p}$ is the learnable filter parameters and $D \in \R^{n \times n}$ is the diagonal degree matrix where the $i$-th diagonal element is defined as $D_i=\sum_{j=1}^n A_{ij}$. 

Using graph convolution as mentioned in Equation~\ref{eqn:existing_graphcnn_eqn}, a set of new node features $X_1=[x_1^1,x_2^1,...,x_n^1]\in\mathbb{R}^{p \times n}$ can be obtained at the output of the first graph convolution layer. Subsequently, using features in $X_1$ as the input node features, the second graph convolution layer produces a set of new node features $X_2$. Similarly, progressive higher level of features $X_3, X_4, \ldots, X_k$ are computed with $k$ layers of graph convolutions. If the graph convolution network has $k$ layers then the final output of the network is, $X_{\text{out}} = X_k$. We use this GNN in this work for ablation studies.

Since the proposed method updates the edge parameters, it is worth mentioning that there is a concept called Dynamic GCN, where the adjacency matrix is also updated after every layer. This was introduced in \cite{EdgeConvolution}. Here they define edge features as $e_{ij} = h_{\theta}(x_i, x_j)$, where $h_{\theta}: \R^{d} \times \R^{d} \rightarrow \R^{d'}$ is a nonlinear function with learnable parameters $\theta$. The output of edge convolution at the i-th vertex is given by $x'_{i} = \Sigma h_{\theta}(x_i, x_j)$. The Adjacency matrix is updated after each layer, by calculating the k-NN neighbours between $\{x'_1, x'_2..., x'_n\}$in the new feature space, thus making the graph dynamic.

However, one may note that

\begin{itemize}
    \item Since the edge features are calculated in this case, there is an apparent similarity with the proposed technique, albeit there is no edge learning, neither are edge features propagated in the next layer.
    \item Though the Dynamic Graph provides global flexibility to the algorithm, this method might not be useful in cases where the graph contains strong local information.
\end{itemize}

We have used this method for a comparative study with the proposed method which is mentioned in section \ref{section:Results}.

\begin{algorithm}
    \caption{Our Approach}
    \label{pseudoCode}
    \begin{algorithmic}[1]
        \Procedure{Train Character Box Predictor}{}
            \State $\textit{M} \gets \text{Initialize UNET with ResNet-50 parameters}$
            \For {$batch = 1,...,B$}
                \State $\textit{Image} \gets \text{Get Batch of Images}$
                \State $\textit{Target} \gets \text{Get Batch of Gaussian Heat-Map char box }$
                \State $Loss_1 \gets MSE(M(Image), Target)$
                \State Update $M$ using Adam optimizer on $Loss_1$
            \EndFor
        \EndProcedure
        
        \Procedure{Train Link Predictor}{}
            \State $\textit{NENET} \gets \text{Initialize with random parameters}$
            \For {$batch = 1,...,B$}
                \State $\textit{Image} \gets \text{Get Batch of Images}$
                \State $\textit{Target CharBoxes} \gets \text{Get Batch of Char Boxes}$
                \State $\textit{Text} \gets \text{Get Batch of Corresponding texts}$
                \State $\textit{Pred CharBoxes} \gets M(Image)$
                \State $\textit{Edges, A} \gets \text{Create graph with k-NN policy}$
                \State $\textit{PredLinks} \gets \textit{NENET}(Pred CharBoxes, Edges, A)$
                \State $\textit{Mapping} \gets \text{Map from Pred to Target CharBoxes}$
                \State $\textit{TargetLinks} \gets GenerateTarget(Mapping, Text)$
                \State $Loss_2 \gets \text{CrossEntropyLoss}(PredLinks, TargetLinks)$
                \State $\text{Update \textit{NENET} using Adam optimizer on $Loss_2$}$
            \EndFor
        \EndProcedure
    \end{algorithmic}
\end{algorithm}

\section{Proposed approach}

For predicting character boxes we use a fully convolutional network architecture based on U-net \cite{UNet}, but instead of predicting affinity boxes to connect adjacent characters as proposed in \cite{CRAFT}, we treat each predicted character as a node and apply graph neural network to predict edge between two characters. A pseudo-code of the proposed approach is tabulated in Algorithm \ref{pseudoCode}.

\subsection{Ground Truth Character Box Generation}

\label{sec:character_detection_method}
Suppose $\{I_1, I_2,..., I_{tr}\}$ be a set of training scene images. Each character of the training scene is annotated using a tight bounding box. It should be noted that considering the possibility of existence of scene characters with different sizes, shapes and styles, the enclosing bounding boxes are created such that they have irregular shapes to conform the shapes of the enclosed characters (please see Figure \ref{fig:CRAFTPipeLine}). Given the fact that neural network can operate more efficiently on a smooth than a sharp distribution space, Gaussian heat maps are generated over the bounding boxes of characters. As similar to \cite{CRAFT}, an initial 2-D isotropic Gaussian heat map is generated which is then transformed into the shape of the corresponding bounding box using a perspective transform computed between the initial map and the bounding box. Such transformed Gaussian heat maps $\{O_1, O_2,..., O_{tr}\}$ are utilized as ground truths for character detection. To train a model for character prediction from unseen samples, We employ a U-Net \cite{UNet} based encoder-decoder architecture which has pre-trained ResNet50 \cite{ResNet} as the encoder. The proposed network has skip connections which allows the aggregation of low-level and high-level features that supplement the model in predicting characters of different sizes. Given the predicted map for character detection corresponding to a sample $I_i$ is $\hat{O}_i$, the following Mean Square Error (MSE) is used as loss to train the proposed character detection model,
\begin{equation}
\label{eqn:loss_char_detect}
    Loss_1 = \sum_{i=1}^{tr}\sum_{(a,b)}\left\lVert \hat{O}_i(a,b)-O_i(a,b)\right\rVert^2
\end{equation}

where $tr$ refers to the number of training samples and $(a, b)$ denotes each pixel. The entire method of ground truth generation for only character detection is the same as CRAFT and is shown in Figure~\ref{fig:CRAFTPipeLine} and the corresponding algorithm is mentioned in detail in Algorithm \ref{pseudoCode}.

\begin{figure}[ht]
    \begin{center}
        \includegraphics[width=\linewidth]{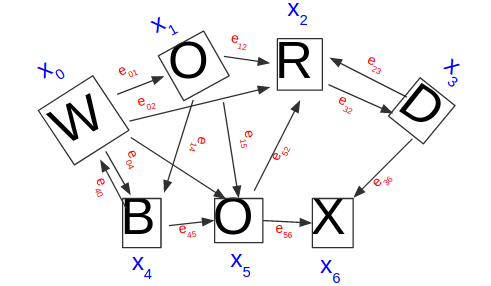}
    \end{center}
    \caption{Creating a graph from predicted character boxes. Note that some edges have not been shown to avoid cluttering.}
    \label{fig:CreatingGraphFromPredictedCharacterBoxes}
\end{figure}

\begin{figure*}[ht]
    \begin{center}
        \includegraphics[width=\textwidth]{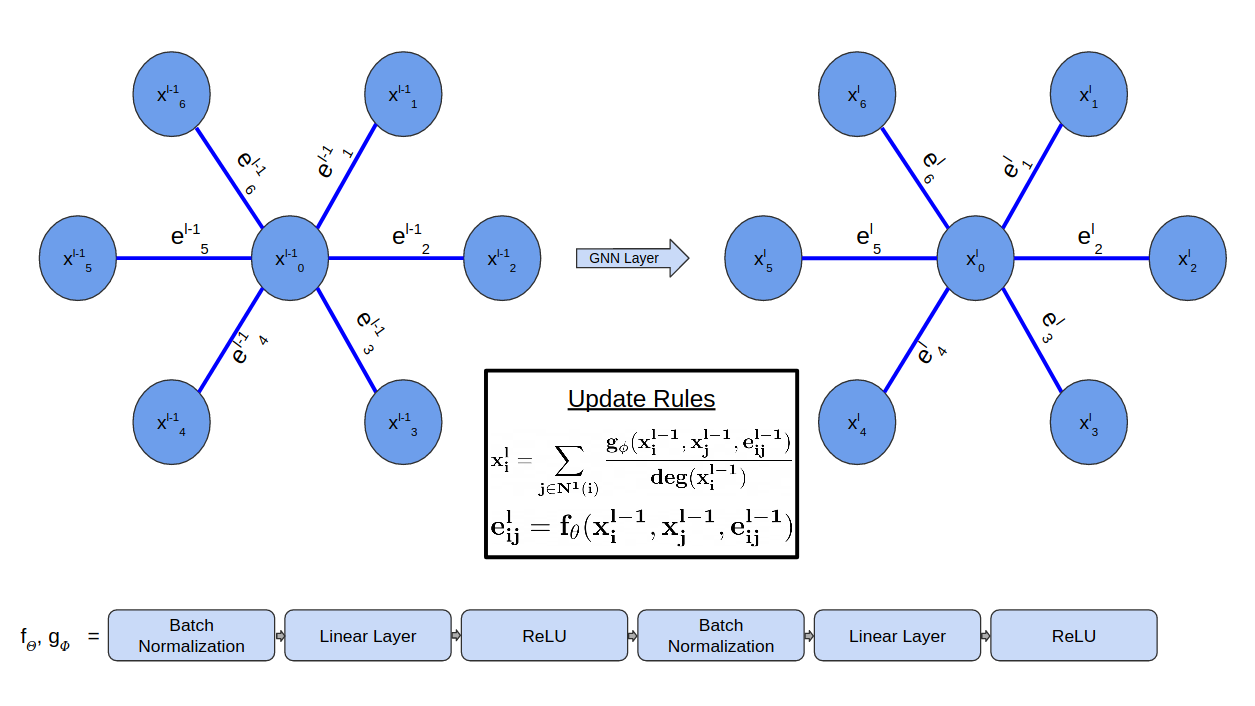}
    \end{center}
    \caption{Graph Update Rule. The notations used are explained in Section \ref{sec:NENET}. Note that it is a directed graph hence $\mathbf{e_{ij} \neq e_{ji}}$.}
    \label{fig:GraphUpdateRule}
\end{figure*}

\subsection{NENET for Edge Prediction}
\label{sec:NENET}
Let a set of characters obtained using the proposed character detection method from a predicted heat map $O$ corresponding to a scene $I$ be denoted by $X=\{x_1, x_2,...,x_n\} \subseteq \R^d $. Each character $x_i, \forall i$ is represented by a $d$ dimensional feature vector. We create a directed graph as $\mathcal{G}=(\mathcal{V}, \mathcal{E})$ over $X$, where each character is a node and each node is connected with it's $k$-nearest neighbours, as shown in Figure \ref{fig:CreatingGraphFromPredictedCharacterBoxes} for illustration. The proposed graph convolution operation is comprised of two steps:(1) \textbf{edge update operation} where at each layer, new edge features are learned for each edge using the features of the connecting nodes and the corresponding edge of the previous layer, (2) \textbf{node update operation} where at each layer, new node features are learned for each node using the features of the connected nodes along with the features of the corresponding connecting edges computed in the previous layer as shown in Figure \ref{fig:GraphUpdateRule}. 
Each node is initialized as $x_i^0=x_i, \forall i$ and each edge is initialized as $e_{ij}^0 = h(x_i, x_j)$ where $h(\cdot)$ along with the graph initialization process is mentioned in detail in section \ref{section:Exp}.

We propose a novel Node and Edge trainable Neural Network called as NENET, and define the operation for edge update at any particular layer ($l$) as follows, 
\begin{equation}
\label{eqn:edge_update}
    e_{ij}^l = f_{\Theta}(x_i^{l-1}, x_j^{l-1}, e_{ij}^{l-1}) 
\end{equation}
where $f_{\Theta}$  could be any parametric differentiable function with parameter $\theta$. For our experiment, we use a 2 layer feed-forward neural network (FFNN).  $e_{ij}^l$ is the edge feature at the layer $l$ between nodes $i$ and $j$. 

The node update operation at any particular layer ($l$) is defined as follows,
\begin{equation}
\label{eqn:node_update}
    x_i^l = \frac{\displaystyle\sum_{j\in \mathcal{N}^1(i)} g_{\phi}(x_i^{l-1}, x_j^{l-1}, e_{ij}^{l-1})}{deg(x_i^{l-1})}
\end{equation}
where $g_{\phi}$  is a 2 layer feed-forward neural network (FFNN) but, it could also be any parametric differentiable function with parameter $\phi$ and $\mathcal{N}^1(\cdot)$ represents first-order neighbors.  $x_i^l$ is the node feature at the layer $l$. The features of the nodes and edges are updated simultaneously at every layer. As there is no implicit order in the node representation, we use the summation operator to ensure that the node update is permutation and degree invariant.

We exploit edge features obtained from the final layer ($L$) of the edge update operation to connect associated characters for a target word detection. To train the proposed NENET, edges $\{e_{ij}\}$ are labeled ($y_{ij}$) as 1 if they connect adjacent characters and as 0 if they do not. It should be noted that an edge that connects adjacent characters is called as a positive edge since it signifies that the connected characters belong to a word. For the similar reason, a particular edge which do not connect characters to make a word is called a negative edge (please see Figure \ref{fig:HowGCNNWorks}). Therefore, to detect a word, each of the associated edges are classified either as positive or negative using ($e_{ij}^L$) obtained from the final layer with the following binary classification operation,
\begin{equation}
\label{eqn:edge_classification}
    \hat{y}_{ij}=Softmax(e_{ij}^L)\quad,\forall e_{ij}\in \mathcal{E}
\end{equation}
We use the following cross-entropy loss to train the proposed NENET model,
\begin{equation}
\label{eqn:NENET_loss}
    Loss_2 = -\sum_{\forall e_{ij}\in \mathcal{E}} y_{ij}\log \hat{y}_{ij}+(1-y_{ij})\log (1-\hat{y}_{ij})
\end{equation}
The proposed NENET consists of $f_{\theta}(\cdot) \text{and} g_{\phi}(\cdot)$ for each layer, and is trained as an end-to-end model using the loss function $Loss_2$.

During testing, given an unseen scene, at first all of the characters are detected using the character detection method as proposed in section \ref{sec:character_detection_method}. Subsequently, a $k$-NN graph is formed over all of the detected characters, and a set of positive and negative edges are predicted using the trained NENET model. Finally, using the predicted positive and negative edges, words are detected, refer to Figure \ref{fig:HowGCNNWorks} for illustration, which is discussed in detail in the following section.

\subsection{Word Detection from a Scene}
Let $\mathcal{E}' \subseteq \mathcal{E}$ be the set of edges which are estimated to be positive. For calculating the word polygons, a new graph $\mathcal{G}'=(\mathcal{V}, \mathcal{E}')$ is created. Using connected components, characters belonging to the same words are isolated and a convex hull is drawn over them as shown in Figure \ref{fig:HowGCNNWorks}. For calculating the word box, we find a minimum area rectangle which tightly encloses a target word.

As can be seen from Table \ref{tab:GCNNComparision}, later in the experimentation section, vanilla GCN performs poorly on the task of link prediction. This is because vanilla GCN treats each edge uniformly and its operations act like a low-pass filter, making it difficult to distinguish adjacent characters. Unlike \cite{EdgeConvolution} who proposed the Dynamic Graph Convolution method using edge convolution, our edge parameters are propagated as well in the next layer. Further, as adjacent characters are always present in the first neighbour of a node, there is no need for a global context when predicting links between them. Hence, as can be seen from Table \ref{tab:GCNNComparision}, having a global flexibility hinders the strong local information present in the graph, lowering the F-score. 

We show that our approach achieves much better results compared to \cite{EdgeConvolution} and other Graph Neural Networks in \hyperref[section:Results]{Results section}.

\section{Experimental Setup}
\label{section:Exp}

\paragraph{Data Set}

As data sets having character level annotation are not readily available, we used SynthText \cite{SynthText} for training and evaluating our method. Current Synthetic datasets are advanced enough that training on synthetic datasets gives good results on real datasets also \cite{SynthText3D, COCO_TS_pixel}.

The SynthText dataset has 858750 images with a total of 7266866 word boxes and 28971487 character boxes. The average height and width of word boxes and character boxes are 28.1 pixels, 75.7 pixels and 23 pixels, 17.4 pixels respectively. It is a huge dataset having a highly variable background, making it ideal for our task. We split the dataset into the ratio of 9:1 for training and testing respectively. We did not use any validation set as we did not do extensive hyper-parameter search and also didn't observe any over-fitting.  

\paragraph{Feature Initialization}
We trained a UNet model having a pre-trained ResNet50 encoder for character prediction. The node features are initialized with the spatial coordinates, width and height of the character box. The edge features are initialized with the relative displacement of the neighboring box centroid with respect to the current box centroid, minimum distance between their corners and relative position(top, bottom, right, left) of the neighboring box with respect to the current box. All the features are normalized between 0 and 1 by the height and width of the image.

\paragraph{Training}

We used 4 GTX1080 Ti for training the model. A batch-size of 12 was used along with a custom decreasing learning-rate regime. To remove the class imbalance and filtering texture-like texts, On-line Hard Negative Mining\cite{OHNM} was applied in the ratio 1:3. To generate the Gaussian heat-map, we used a window size of 50 pixels with a standard deviation of 18.5 pixels. The graph was calculated using 4 nearest neighbour approach. The architecture of the GNN contained 2 graph layers. This was calculated using experiments for hyper-parameter selection and matches the intuition that most of the information for link prediction is available in the proximity of the nodes. The input node features were 16-dimension, input edge features were 6-dimension. The output edge features were 2-dimension to train using cross-entropy loss. The output node features were irrelevant for link prediction but intermediate node outputs affect edge prediction.

\paragraph{Mapping Predicted to Target Character Boxes}

While training NENET, instead of training on oracle character boxes, we used the character boxes estimated by the UNet network. For generating the link targets, we first found a mapping from the predicted to target character boxes. This is done by thresholding the IOU of predicted and target character boxes at 0.25. 

As can be seen in Figure \ref{fig:HowGCNNWorks}, links which connect the next character belonging to the same word are taken as positive and all other links are taken as negative. If the next character does not have any corresponding mapping in the target character box, then the link to the character after the adjacent next character is taken as positive. All predicted characters which do not have any mapping to the target characters have all their links as negative.

\paragraph{Using Vanilla GCN}

\begin{figure*}[!ht]
    \begin{center}
        \includegraphics[width=0.41\linewidth]{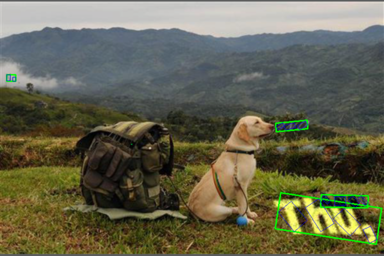}
        \includegraphics[width=0.41\linewidth]{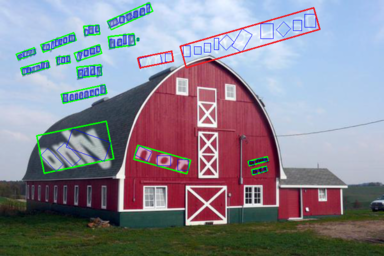}
        \includegraphics[width=0.41\linewidth]{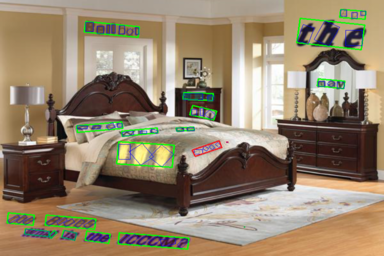}
        \includegraphics[width=0.41\linewidth]{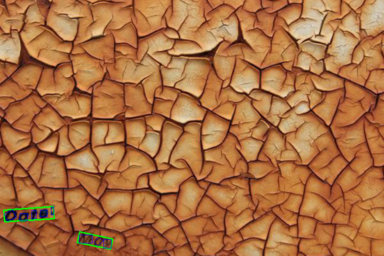}
        \includegraphics[width=0.41\linewidth]{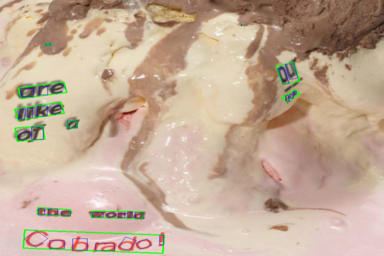}
        \includegraphics[width=0.41\linewidth]{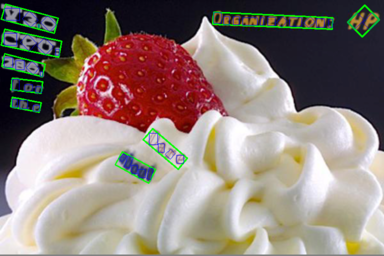}
        \includegraphics[width=0.41\linewidth]{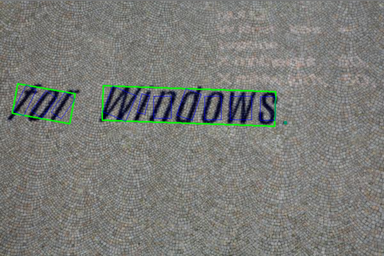}
        \includegraphics[width=0.41\linewidth]{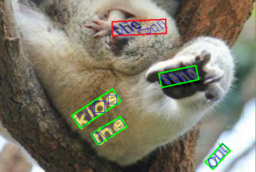}
    \end{center}
    \caption{Qualitative display of results of predicted words from the given character boxes.Color legends are: predicted word box (green for correct, red for incorrect) and character box (blue).}
    \label{fig:results}
\end{figure*}

As Vanilla GCN does not predict edge features, we had to modify the approach to link prediction. First using vanilla GCN, we extract $n$ dimensional node features individually from $x_i, x_j$. Let us denote them by $y_i, y_j,$ respectively. To predict link between them, we applied a feed-forward neural network on $y_i, y_j$, which gives a binary output denoting presence or absence of links. As can be seen from Table \ref{tab:GCNNComparision}, this method performs poorly for reasons mentioned in the Section \ref{section:Results}.

\begin{figure*}[ht]
    \begin{center}
        \includegraphics[width=0.4\linewidth]{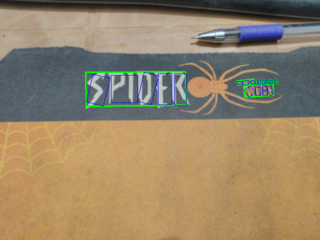}
        \includegraphics[width=0.4\linewidth]{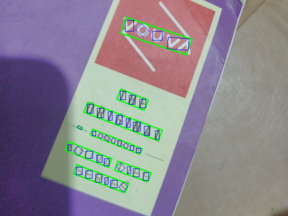}
        \caption{Qualitative cross domain performance evaluation of the proposed model. It is trained on synthetic dataset but applied on real images.}
    \end{center}
    \label{fig:realResults}
\end{figure*}

\paragraph{Using Dynamic GCN}

The experimental setup for Dynamic GCN was similar to NENET. Three major differences however were that the Adjacency matrix was calculated after every layer for making the graph dynamic, edge features were not propagated between layers and instead of 2 hidden layers, the number of layers were increased to 4 which gave slightly better performance. For predicting links in the final layer, we used the edge features generated from the final layer. The results are mentioned in Table \ref{tab:GCNNComparision}.

\paragraph{Weak-Supervision for natural scene text detection}

\cite{CRAFT} used weak-supervision to transfer parameters learnt from synthetic dataset having character box annotation to real datasets which did not have character box annotation. Unfortunately, we were not able to reproduce the results mentioned in \cite{CRAFT} due to lack of adequate information on training hyper-parameters. The model we were able to reproduce by weak supervision produced smudgy character predictions, hence it was not possible to get separate character boxes for training NENET on the real dataset.

\section{Results}
\label{section:Results}

\begin{table}[ht]
    \caption{Comparison of F-score of different GNN methods with NENET on the SynthText data set. Dynamic GCN refers to \cite{EdgeConvolution}, Vanilla GCN refers \cite{kipf2016semi}.}
    \label{tab:GCNNComparision}
    \centering
    \begin{tabular}{c| c}
         \hline
         Method & F-Score\\
         \hline
         CRAFT & 0.622\\
         Dynamic GCN & 0.542\\
         Vanilla GCN & 0.080\\
         NENET w/o Edge Propagation & 0.604\\
         NENET & \textbf{0.655}\\
         \hline
    \end{tabular}
\end{table}

We summarize the effectiveness of our method by comparing it with other methods on the SynthText dataset in Table \ref{tab:GCNNComparision}. We train and evaluate all methods on synthetic data set. We also compare our graph neural network method with \cite{EdgeConvolution, kipf2016semi}.

As can be noted from Table \ref{tab:GCNNComparision}, vanilla GCNN gives very poor results. We believe this to be the case because vanilla GCN has a low pass filter operation, which diffuses the information, making it harder for deeper layers to predict links. Dynamic GCN leverages similarity in the global context to group nodes. While this has interesting applications, there is not much difference in features between characters in the global context. Even the existing CRAFT \cite{CRAFT} algorithm performs better than the dynamic GCN. The proposed method outperforms all other methods. We also perform an ablation study to validate the use of edge propagation. For the ablation study, we fixed the edges with the initial features keeping the model parameters nearly the same, and trained them independently. The low score of the model without edge propagation confirms the advantage of having edge features that are updated and propagated through the layers.

In figure \ref{fig:results}, we show the sample results of text detection qualitatively using the proposed technique. One can see that the method is able to link the character boxes into words quite efficiently. Some of the words are very short, while several of them are long. But they could be identified properly. Further the gap between character boxes are also quite variable. Some of the words deviate from horizontal orientation. Also detected words like "about" and "Cobrado!" have been written along a curved trajectory. It may also be seen from these images that there is a wide variability in the scene texture. Notwithstanding the above variations, the proposed algorithm has successfully linked the character boxes in most cases. Needless to say that the proposed method does miss a few words and those have been shown in red in the figures. Most of the errors involve combining two or more closely spaced words into a single one.

The current limitation of our approach is that it can be applied to datasets that have prior character level bounding box annotations. This is required for training purposes. Without having access to such training datasets, we are unable to demonstrate the performance of the proposed technique on large scale real data. Since such datasets are not readily available, we now show some sample results on real images in Figure \ref{fig:realResults}. Without doing any retraining or fine-tuning of parameters we simply apply the proposed technique trained on SynthText data on arbitrary scenes. The qualitative results are very promising as can be seen from these figures. Only the word 'precision' in 'high precision mice' has been over-segmented into multiple words. Fine-tuning on real data sets having character level annotations should further improve the predictions. 

\section{Conclusion}
In this paper we propose a novel approach of Graph Neural Network which uses edge propagation in conjugation with node propagation to link predicted characters. In spite of a high variance of word lengths, the proposed method can connect characters into words quite accurately. Currently we are working towards combining character recognition engine to the proposed NENET architecture so that we can leverage on the use of lexicon to further improve link prediction. 

\bibliographystyle{IEEEbib}
\bibliography{main}

\end{document}